\documentclass[10pt,twocolumn,letterpaper]{article}
\usepackage{iccv}
\usepackage{times}
\usepackage{epsfig}
\usepackage{graphicx}
\usepackage{amsmath}
\usepackage{amssymb}

\usepackage{tikz}
\usepackage{xcolor}
\usepackage{xspace}
\usepackage{tabularx}
\usepackage{booktabs} 
\usepackage{mathtools}
\usepackage{caption}
\usepackage{subfigure}
\usepackage{algpseudocode,algorithm}

\usepackage[pagebackref=true,breaklinks=true,letterpaper=true,colorlinks,bookmarks=false]{hyperref}

\iccvfinalcopy 

\def\iccvPaperID{1874} 


\makeatletter
\newcommand{\todo}[1][]{\@latex@warning{TODO #1}}
\makeatother

\newlength\figureheight
\newlength\figurewidth

\newif\ifshowplots
\showplotstrue 

\definecolor{dmyellow}{HTML}{F8E46C}
\definecolor{dmlightblue}{HTML}{7DC3D2}
\definecolor{dmred}{HTML}{EF5A58}
\definecolor{dmpurple}{HTML}{DA81B3}
\definecolor{dmgreen}{HTML}{BCD26F}

\usetikzlibrary{shapes, arrows, fit}
\usetikzlibrary{positioning,shadows}
\usetikzlibrary{decorations.pathreplacing}
\definecolor{inodefill}{rgb} {0.00,0.53,0.88}
\definecolor{inodedraw}{rgb} {1.00,1.00,1.00}
\definecolor{pnodefill}{rgb} {1.00,1.00,1.00}
\definecolor{onodedraw}{rgb} {0.00,0.00,0.00}
\definecolor{onodefill}{rgb} {1.00,1.00,1.00}
\definecolor{pnodedraw}{rgb} {0.00,0.00,0.00}
\definecolor{mnodefill}{rgb} {0.00,0.53,0.88}

\usetikzlibrary{backgrounds}
\usetikzlibrary{calc}
\usetikzlibrary{external}
\usetikzlibrary{fit}
\usetikzlibrary{positioning}
\usetikzlibrary{shapes}
\tikzexternaldisable

\usepackage[symbol]{footmisc}

\begin{document}
	
	\title{Learning to Paint With Model-based Deep Reinforcement Learning}
	
	\author{Zhewei Huang$^{1,2}$
		~~~~
		Wen Heng$^{1}$
		~~~~
		Shuchang Zhou$^{1}$\\
		$^{1}$Megvii Inc~~~~$^{2}$Peking University\\
		{\tt\small \{huangzhewei, hengwen, zsc\}@megvii.com}\\
	}
	
	\maketitle
	
	\begin{abstract}
		We show how to teach machines to paint like human painters, who can use a small number of strokes to create fantastic paintings. By employing a neural renderer in model-based Deep Reinforcement Learning (DRL), our agents learn to determine the position and color of each stroke and make long-term plans to decompose texture-rich images into strokes. Experiments demonstrate that excellent visual effects can be achieved using hundreds of strokes. The training process does not require the experience of human painters or stroke tracking data. The code is available at \url{https://github.com/hzwer/ICCV2019-LearningToPaint}.
	\end{abstract}
	
	\section{Introduction}
    Painting, being an important form of visual art, symbolizes human wisdom and creativity. In recent centuries, artists have used a diverse array of tools to create their masterpieces. But it's hard for people to master this skill without spending a large amount of time in proper training. Therefore, teaching machines to paint is a challenging task and helps to shed light on the mystery of painting. Furthermore, the study of this topic can help us build painting assistant tools.
    
    We train an artificial intelligence painting agent that can paint strokes on a canvas in sequence to generate a painting that resembles a given image. Neural networks are used to produce parameters that control the position, shape, color, and transparency of strokes. Previous works have studied teaching machines to learn painting-related skills, such as sketching~\cite{ha2017neural,chen2017sketch,song2018learning}, doodling~\cite{zhou2018learning} and writing characters~\cite{zheng2018strokenet}. In contrast, we aim to teach machines to handle more complex tasks, such as painting portraits of humans and natural scenes in the real world, which have rich textures and complex structural compositions.
    
    \begin{figure}[t]%
    \centering
    \begin{minipage}[t]{0.158\linewidth}
    \vspace{0pt}
    \includegraphics[width=1\linewidth]{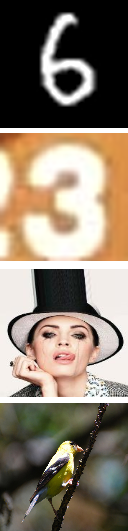}
    \end{minipage}~
    \begin{minipage}[t]{0.79\linewidth}
    \vspace{0pt}
    \includegraphics[width=1\linewidth]{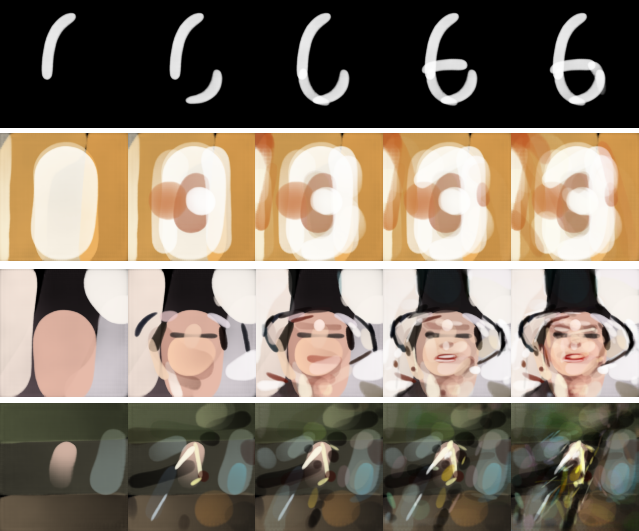}
    \end{minipage}
        \caption{\textbf{The painting process}. The first column shows the target images. Our agent tends to draw in a coarse-to-fine manner.
}
\vspace{-0.2cm} 
    \end{figure}
    
        \begin{figure*}[ht]%
    \centering
    \begin{minipage}[b]{0.44\linewidth}
    \centering
    \subfigure[]{\includegraphics[width=1\textwidth]{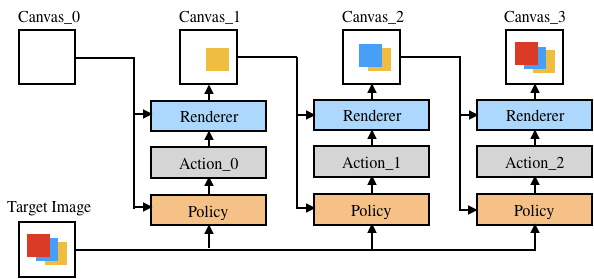}}
    \end{minipage}
    \begin{minipage}[b]{0.54\linewidth}
    \centering
    \subfigure[]{\includegraphics[width=0.9\textwidth]{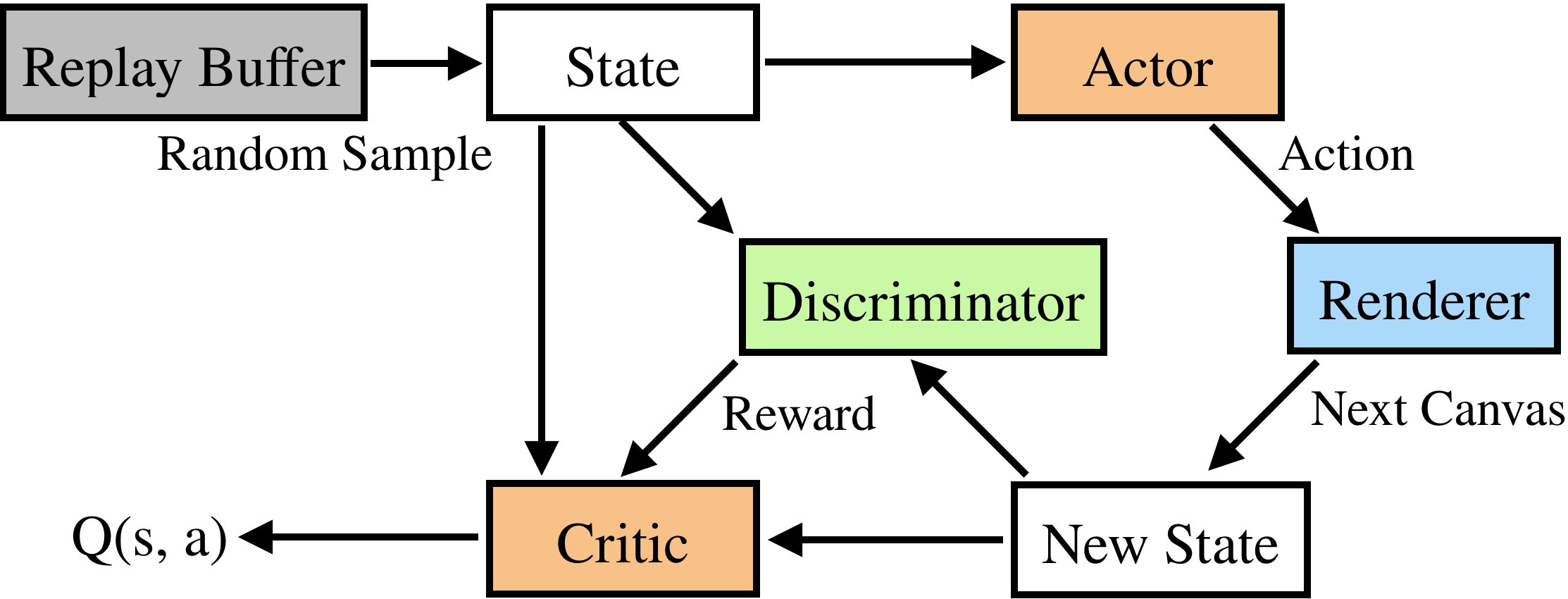}}\\
    \end{minipage}
    \vspace{-0.2cm} 
        \caption{\textbf{The overall architecture}. (a) At the inference stage, the actor outputs a set of stroke parameters based on the canvas status and target image at each step. The renderer then renders the stroke on the canvas accordingly. (b) At the training stage, the actor is trained with assistants of an adversarial discriminator and a critic.  The reward is given by the discriminator at each step, and the training samples are randomly sampled from the replay buffer.}
        \label{fig:pipeline}
    \end{figure*}
    
        We address three challenges for training an agent to paint real-world images. First, painting like humans requires an agent to have the ability to decompose a given target image into an ordered sequence of strokes. The agent needs to parse the target image visually, understand the current status of the canvas, and have foresightful plans about future strokes. To achieve this planning, one method is to give the supervised loss for stroke decomposition at each step, as in~\cite{ha2017neural}. However, such a method require ground truth stroke decomposition, which is hard to define. Also, texture-rich image painting usually requires hundreds of strokes to generate a painting that resembles a target image, which is tens of times more than doodling, sketching or character writing require and increases the difficulty of planning. To handle the ill-definedness of the problem, and the long-term planning challenge, we propose using reinforcement learning (RL) to train the agent, because RL can maximize the cumulative rewards of a whole painting process rather than minimizing supervised loss at each step. Experiments show that an RL agent can build plans for stroke decomposition with hundreds of steps. Moreover, we apply the adversarial training strategy~\cite{Goodfellow2014gan} to improve the pixel-level quality of the generated images, as the strategy has proved effective in other image generation tasks~\cite{Ledig2016Photo}.
    
    Second, we design continuous stroke parameter space, including stroke location, color, and transparency, to improve the painting quality. Previous works~\cite{ha2017neural,zhou2018learning,ganin2018synthesizing} design discrete stroke parameter spaces and each parameter has only a limited number of choices, which fall short for texture-rich paintings. Instead, we adopt the Deep Deterministic Policy Gradient (DDPG)~\cite{lillicrap2015continuous} which copes well with the continuous action space of the agent.
    
    Third, we build an efficient differentiable neural renderer that can simulate painting of hundreds of strokes on the canvas. Most previous works~\cite{ha2017neural,zhou2018learning,ganin2018synthesizing} paint by interacting with undifferentiable painting simulation environments, which are good as renders but fail to provide detailed feedback about the generated images. Instead, we train a neural network that directly maps stroke parameters to stroke paintings. The renderer can also be adapted to different stroke designs like triangle and circles by changing the generation patterns. Moreover, the differential renderer can be combined with DDPG into a single model-based DRL that can be trained in an end-to-end fashion, which significantly boosts both the painting quality and convergence speed.
    
    In summary, our contributions are three-fold:
    \begin{itemize}
        \item We approach the painting task with the model-based DRL algorithm and build agents that decompose the target image into hundreds of strokes in sequence which can recreate a painting on canvas. 
        
        \item We build differentiable neural renderers for efficient painting and flexible support of different stroke designs, \eg B{\'e}zier curve, triangle, and circle. The neural renderer contributes to the painting quality by allowing training model-based DRL agent in an end-to-end fashion. 
        
        \item Experiments show that the proposed painting agent can handle multiple types of target images well, including handwritten digits, streetview house numbers, human portraits, and natural scene images.
    \end{itemize}
	\begin{figure*}[h]
  \centering
  \begin{minipage}[b]{0.18\linewidth}
  \centering
  \subfigure[MNIST~\cite{lecun1998mnist}]{\includegraphics[width=0.995\textwidth]{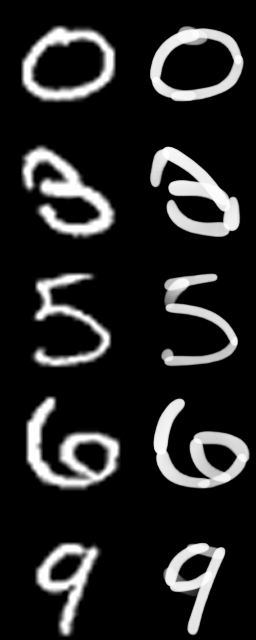}}
  \end{minipage}
  \begin{minipage}[b]{0.18\linewidth}
  \centering
  \subfigure[SVHN~\cite{netzer2011reading}]{\includegraphics[width=0.995\textwidth]{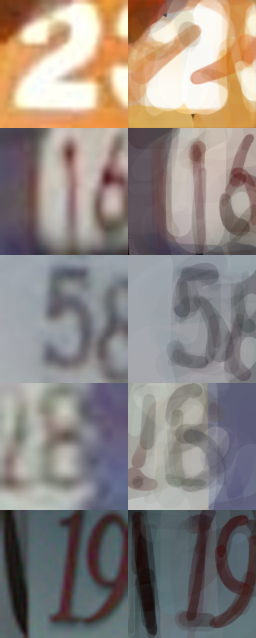}}\\
  \end{minipage}
  \begin{minipage}[b]{0.3\linewidth}
  \centering
  \subfigure[CelebA~\cite{liu2015faceattributes}]{\includegraphics[width=0.995\textwidth]{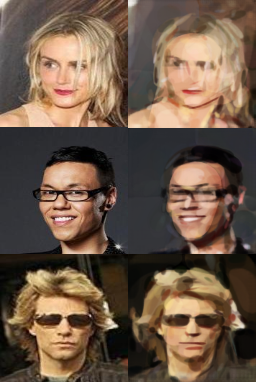}}\\
  \end{minipage}
  \begin{minipage}[b]{0.3\linewidth}
  \centering
  \subfigure[ImageNet~\cite{ILSVRC15}]{\includegraphics[width=0.995\textwidth]{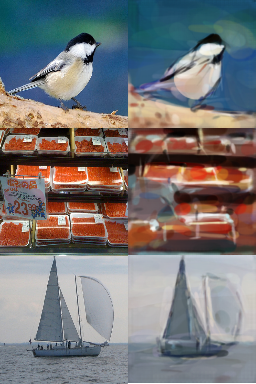}}\\
  \end{minipage}
   \caption{The painting results on multiple datasets. The stroke numbers of the paintings are 5, 40, 200 and 400 for MNIST, SVHN, CelebA and ImageNet respectively.}
   \vspace{-0.2cm}
        \label{fig:4dataset}
\end{figure*}

\section{Related work}
    Stroke-based rendering (SBR) is a method of non-photorealistic imagery that recreates images by placing discrete drawing elements such as paint strokes or stipples~\cite{hertzmann2003survey} on canvas. Most SBR algorithms solve the stroke decomposition problem by Greedy Search on every single step or require user interaction. Haeberli~\etal~\cite{haeberli1990paint} propose a semiautomatic method which requires the user to set parameters to control the shape of the strokes and select the positions for each stroke. Litwinowicz~\etal~\cite{litwinowicz1997processing} propose a single-layer painter-like rendering which places the brush strokes on a grid in the image plane, with randomly perturbed positions. Some work also studies the effects of using different stroke designs~\cite{hertzmann1998painterly} and the related problem of generating animations from video~\cite{lin2010painterly}.
    
    Recent works use RL to improve the stroke decomposition of images.
    SPIRAL~\cite{ganin2018synthesizing} is an adversarially trained DRL agent that learn structures in images, but fails to recover the details of human portraits. StrokeNet~\cite{zheng2018strokenet} combines differentiable renderer and recurrent neural network (RNN) to train agents to paint but fails to generalize on color images. Doodle-SDQ~\cite{zhou2018learning} trains the agents to emulate human doodling with DQN. Earlier, Sketch-RNN~\cite{ha2017neural} uses sequential datasets to achieve good results in sketch drawings. Artist Agent~\cite{xie2013artist} explores using RL for the automatic generation of a single brush stroke. 
     
     %
    
	\section{Painting Agent}
\subsection{Overview}
The goal of the painting agent is decomposing the given target image into strokes that can recreate the image on the canvas. To imitate the painting process of humans, the agent is designed to predict the next stroke based on observing the current state of the canvas and the target image. However, the stroke at each step needs to be well compatible with previous strokes and future strokes to reduce the number of strokes for finishing the painting. We postulate that the agent should maximize the cumulative rewards after finishing the given number of strokes, rather than the gain of current stroke. To achieve this delayed-reward design, we employ a DRL framework, with the diagrams for the overall architecture shown in Figure~\ref{fig:pipeline}.

In the framework, we model the painting process as a sequential decision-making task, which is described in Section~\ref{subsec:model}. And to build the feedback mechanism, we use a neural renderer to help generate detailed rewards for training the agent, which is described in Section~\ref{subsec:learning}. 

\subsection{The Model}
\label{subsec:model}

Given a target image $I$ and an empty canvas $C_0$, the agent aims to find a stroke sequence $(a_0, a_1,...,a_{n-1})$, where rendering $a_t$ on $C_t$ can get $C_{t+1}$. After rendering these strokes in sequence, we get the final painting $C_n$, which should be visually similar to $I$ as much as possible. We model this task as a Markov Decision Process with a state space $\mathcal{S}$, an action space $\mathcal{A}$, a transition function $\operatorname{trans}(s_t, a_t)$ and a reward function $r(s_t, a_t)$. The details of these components are specified next.
    
    \textbf{State and Transition Function} The state space is constructed by all possible information that the agent can observe in the environment. We separate a state into three parts: states of the canvas, the target image, and the step number. Formally, $s_t = (C_t, I, t)$. $C_t$ and $I$ are bitmaps and the step number $t$ acts as additional information to instruct the agent the remaining number of steps. The transition function , $s_{t+1} = \operatorname{trans}(s_t, a_t)$ gives the transition process between states, which is implemented by painting a stroke on the current canvas.
    
    \textbf{Action} An action $a_t$ of the painting agent is a set of parameters that control the position, shape, color and transparency of the stroke that would be painted at step $t$. We define the behavior of an agent as a policy function $\pi$ that maps states to deterministic actions, i.e. $\pi \colon \mathcal{S} \to \mathcal{A}$. At step $t$, the agent observes state $s_t$ before predicting the parameters of the next stroke $a_t$. The state evolves based on the transition function $s_{t+1} = \operatorname{trans}(s_t, a_t)$, which runs for $n$ steps. 
    
    \textbf{Reward} 
    Selecting a suitable metric to measure the difference between the current canvas and the target image is found to be crucial for training a painting agent.
    The reward is designed as follows,
    \begin{equation}
    \label{eq:reward}
    r(s_t, a_t)= L_{t}-L_{t+1}
    \end{equation}
    where $r(s_t, a_t)$ is the reward at step $t$, $L_{t}$ is the measured loss between $I$ and the $C_{t}$ and $L_{t+1}$ is the measured loss between $I$ and the $C_{t+1}$. In this work, $L$ is formulated as the discriminator score that is defined in Section~\ref{sec:wgan_reward}.
    
    To make the final canvas resemble the target image, the agent should be driven to maximize the cumulative rewards in the whole episode. At each step, the objective of the agent is to maximize the sum of discounted future reward $R_t = \sum_{i = t}^{T} \gamma^{(i - t)} r(s_i, a_i) $ with a discounting factor $\gamma \in [0, 1] $. 

\subsection{Learning}
\label{subsec:learning}
    In this section, we introduce how to train the agent using the model-based DDPG algorithm. 
    
\begin{figure}[h]
  \centering
  \begin{minipage}[b]{0.45\linewidth}
  \centering
  \subfigure[Original DDPG]{\includegraphics[width=1\textwidth]{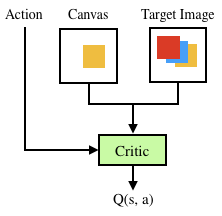}}
  \end{minipage}
  \begin{minipage}[b]{0.53\linewidth}
  \centering
  \subfigure[Model-based DDPG]{\includegraphics[width=1\textwidth]{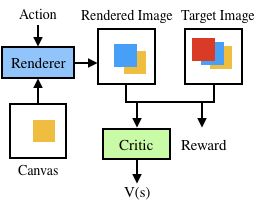}}\\
  \end{minipage}
  \vspace{-0.2cm} 
  \caption{In the original DDPG, critic needs to learn to model the environment implicitly. In the model-based DDPG, the environment is explicitly modeled through a neural renderer, which helps to train an agent efficiently.}
    \label{fig:DDPG}
    \vspace{-0.3cm} 
\end{figure}

    \subsubsection{Model-based DDPG}
    We first describe the original DDPG, then introduce building model-based DDPG for efficient agent training. 
    
    As we use continuous parameters for strokes, the action space in the painting task is continuous and of high dimensions. Discretizing the action space to adapt some DRL methods, such as DQN and PG, will lose the precision of stroke representation and require many efforts in manual structure design to cope with the explosion of parameter combinations in discrete space. In contrast, DPG~\cite{silver2014deterministic} uses deterministic policy to resolve the difficulties caused by high-dimensional continuous action space, and DDPG is its variant using Neural Networks.
    
    In the original DDPG, there are two networks: the actor $\pi(s)$ and critic $Q(s, a)$. The actor models a policy $\pi$ that maps a state $s_t$ to action $a_t$. The critic estimates the expected reward for the agent taking action $a_t$ at state $s_t$, which is trained using Bellman equation~(\ref{eq:bellman}) as in Q-learning~\cite{watkins1992q} and the data is sampled from an experience replay buffer:
    
    \begin{equation}
    \label{eq:bellman}
        Q(s_t, a_t) = r(s_t, a_t) + \gamma Q(s_{t+1},\pi({s_{t+1}}))
    \end{equation}.
    
    Here $r(s_t, a_t)$ is a reward given by the environment when performing action $a_t$ at state $s_t$. The actor $\pi(s_t)$ is trained to maximize the critic's estimated $Q(s_t, \pi(s_t))$. In other words, the actor decides a stroke for each state. Based on the current canvas and the target image, the critic predicts an expected reward for the stroke. The critic is optimized to estimate more accurate expected rewards.
    
    We cannot train a good-performance painting agent using original DDPG because it's hard for the agent to model the complex environment well that is composed of any types of real-world images during learning. The World Model~\cite{ha2018world} is a method to make agent understand the environments effectively. Similarly, we design a neural renderer so that the agent can observe a modeled environment. Then it can explore the environment and improve its policy efficiently. We term the DDPG with the actor that can get access to the gradients from environments as model-based DDPG. The difference between the two algorithms is visually shown in Figure~\ref{fig:DDPG}. 
    
    The optimization of the agent using the model-based DDPG is different from that using the original DDPG. At step $t$, the critic takes $s_{t+1}$ as input rather than both of $s_{t}$ and $a_{t}$. The critic still predicts the expected reward for the state but no longer includes the reward caused by the current action. The new expected reward is a value function $V(s_t)$ trained using discounted reward:
    \begin{equation}
        V(s_t) = r(s_{t}, a_{t})+ \gamma V(s_{t+1})
    \end{equation}
    Here $r(s_t, a_t)$ is the reward when performing action $a_t$ based on $s_{t}$. The actor $\pi(s_t)$ is trained to maximize $r(s_t, \pi(s_t)) + V(\operatorname{trans}(s_t, \pi(s_t)))$. The transition function $s_{t+1} = \operatorname{trans}(s_t, a_t)$ is the differentiable renderer.

    \subsubsection{Action Bundle}
    Frame Skip~\cite{braylan2015frame} is a powerful trick for many RL tasks, by restricting the agent to only observe the environment and acts once every $k$ frames rather than one frame. The trick makes the agents have a better ability to learn associations between more temporally distant states and actions. The agent predicts one action and reuse it at the next $k-1$ frames instead and achieves better performance with less computation cost.
    
    Inspired by this trick, we propose using Action Bundle that the agent predicts $k$ strokes at each step and the renderer renders these strokes in order. This practice encourages the exploration of the action space and action combinations. The renderer can render $k$ strokes simultaneously to greatly speed up the painting process. 
    
    We experimentally find that setting $k=5$ is a good choice that significantly improves the performance and the learning speed. It's worth noting that we modify the reward discount factor from $\gamma$ to $\gamma^k$ to keep consistency.

    \subsubsection{WGAN Reward}
    \label{sec:wgan_reward}
\begin{figure}[h]%
    \centering \includegraphics[width=0.35\textwidth]{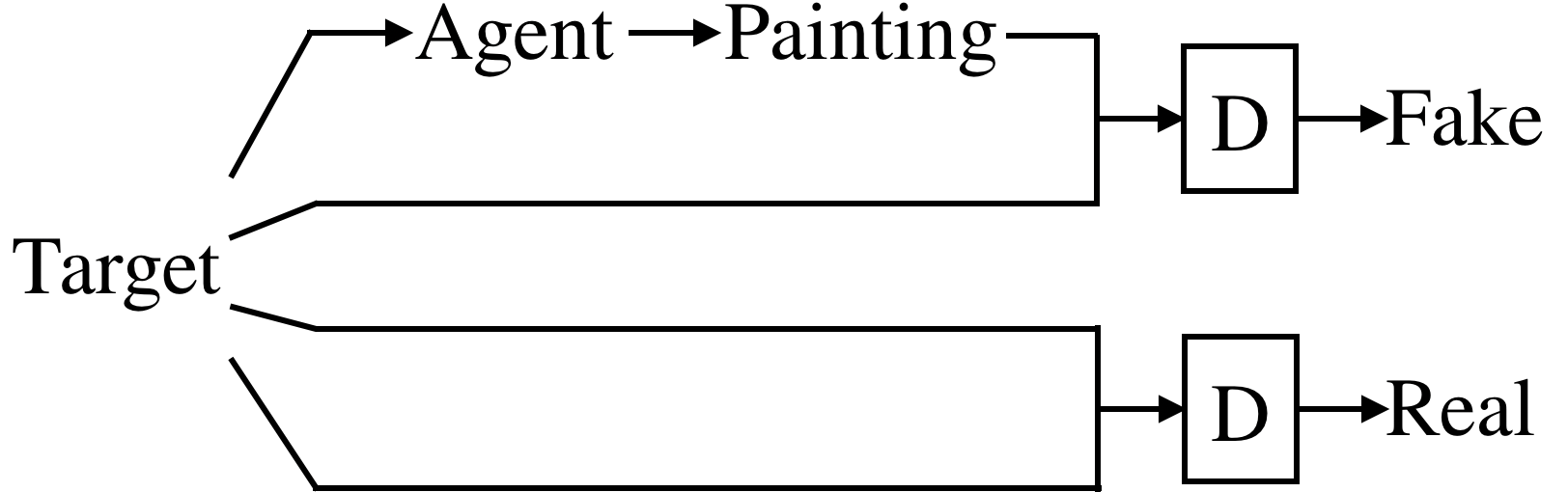}
    \caption{Training of the Discriminator}
    \label{fig:wgan}
    \vspace{-0.2cm}
    \end{figure}

    GAN has been widely used as a particular loss function in transfer learning, text model and image restoration~\cite{ledig2017photo, yeh2017semantic}, because of its great ability in measuring the distribution distance between the generated data and the target data. Wasserstein GAN (WGAN)~\cite{arjovsky2017wasserstein} is an improved version of the original GAN that uses the \emph{Wasserstein-l} distance, also known as \emph{Earth-Mover} distance. The objective of the discriminator in WGAN is defined as 
    \begin{equation}
            \max\limits_{D} \mathbb{E}_{y\sim \mu} [D(y)]-\mathbb{E}_{x\sim \nu}[D(x)]
    \end{equation}
    where $D$ denotes the discriminator, $\nu$ and $\mu$ are the fake samples and real samples distribution. The conditional GAN training schema ~\cite{isola2017image} is used, where fake samples are pairs of a painting and its target; and real samples are two same target images as shown in Figure~\ref{fig:wgan}. The prerequisite of the above objective is that $D$ should be under the constraints of 1-Lipschitz. To achieve the constraint, we use WGAN with gradient penalty (WGAN-GP)~\cite{gulrajani2017improved}.
    
    We want to reduce the differences between paintings and target images as much as possible. To achieve this, we set the difference of $D$ scores from $s_t$ to $s_{t+1}$ using equation~(\ref{eq:reward}) as the reward for guiding the learning of the actor. In experiments, we find rewards derived from $D$ scores is better than $L_2$ distance.
    
    
    
\subsection{Network Architectures}
    
   \begin{figure}[h]
  \centering
  \begin{minipage}[b]{0.52\linewidth}
  \centering
  \subfigure[The actor and critic]{\includegraphics[width=1\textwidth]{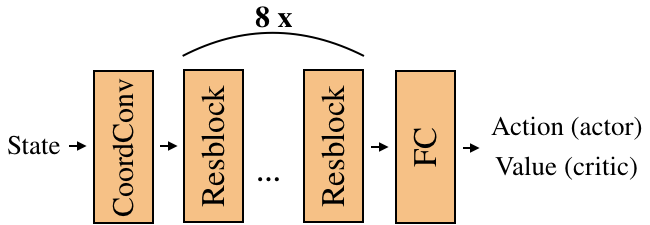}}
  \end{minipage}
  \begin{minipage}[b]{0.465\linewidth}
  \centering
  \subfigure[The discriminator]{\includegraphics[width=1\textwidth]{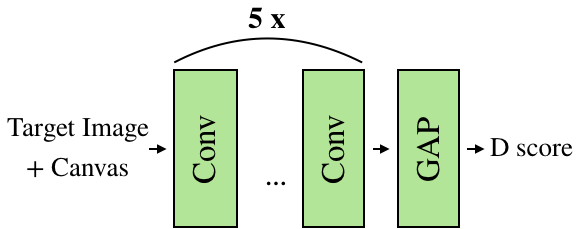}}\\
  \end{minipage}
  \centering
  \begin{minipage}[t]{0.65\linewidth}
  \centering
  \subfigure[The neural renderer]{\includegraphics[width=1\textwidth]{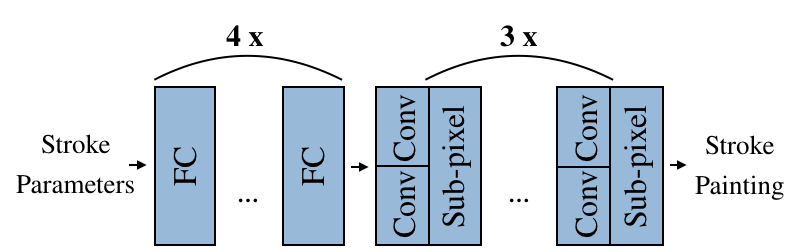}}\\
  \end{minipage}
  \vspace{-0.1cm} 
  \caption{\textbf{Network architectures.} \texttt{FC} refers to a fully-connected layer, \texttt{Conv} refers to a convolution layer, and \texttt{GAP} refers to a global average pooling layer. The actor and the critic use the same structure except for the last FC layers that have different output dimensions.}
    \label{fig:net_arch}
      \vspace{-0.1cm} 
\end{figure}
    
    Due to the high variability and complexity of real-world images, we use residual structures similar to ResNet-18~\cite{he2016deep} as the feature extractor in the actor and the critic. The actor works well with Batch Normalization (BN)~\cite{ioffe2015batch}, but BN can not speed up the critic learning significantly. We use WN~\cite{salimans2016weight} with Translated ReLU (TReLU)~\cite{xiang2017effects} on the critic to stabilize our learning. In addition, we use CoordConv~\cite{liu2018intriguing} as the first layer in the actor and the critic. For the discriminator, we use a network architecture similar to PatchGAN~\cite{isola2017image}, and with WN and TReLU. The network architectures of the actor, critic and discriminator are shown in Figure~\ref{fig:net_arch} (a) and (b).
    
    Following the original DDPG paper, we use the soft target network which creates a copy for the actor and critic and updating their parameters by having them slowly track the learned networks. We also apply this trick on the discriminator to improve its training stability.
    

	\begin{figure*}[htbp]
  \centering
  \begin{minipage}[t]{0.2\linewidth}
  \centering
  \subfigure[]{\includegraphics[width=0.98\textwidth]{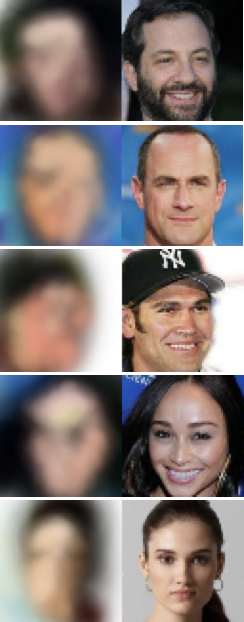}}
  \end{minipage}
  \begin{minipage}[t]{0.125\linewidth}
  \centering
  \subfigure[]{\includegraphics[width=0.98\textwidth]{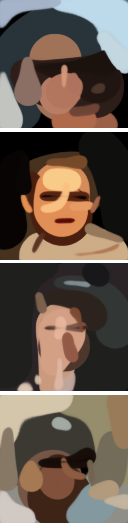}}
  \end{minipage}
  \begin{minipage}[t]{0.125\linewidth}
  \centering
  \subfigure[]{\includegraphics[width=0.98\textwidth]{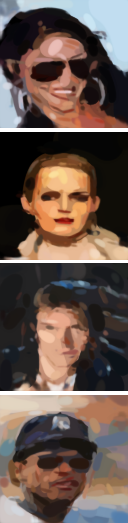}}
  \end{minipage}
  \begin{minipage}[t]{0.125\linewidth}
  \centering
  \subfigure[]{\includegraphics[width=0.98\textwidth]{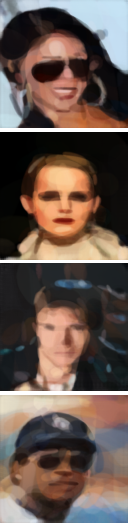}}
  \end{minipage}
  \begin{minipage}[t]{0.125\linewidth}
  \centering
  \subfigure[]{\includegraphics[width=0.98\textwidth]{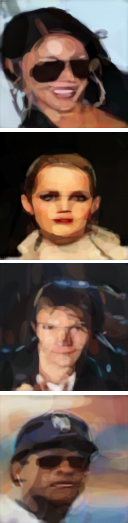}}
  \end{minipage}
  \begin{minipage}[t]{0.125\textwidth}
  \centering
  \subfigure[]{\includegraphics[width=0.98\textwidth]{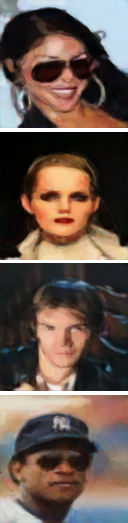}}
  \end{minipage}
  \begin{minipage}[t]{0.125\textwidth}
  \centering
  \subfigure[]{\includegraphics[width=0.98\textwidth]{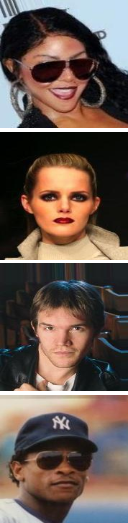}}
  \end{minipage}
  \vspace{-0.2cm} 
  \caption{CelebA paintings under different settings. (a) The paintings of SPIRAL with 20 strokes~\cite{ganin2018synthesizing} (b) Ours with 20 opaque strokes (c) Ours with 200 opaque strokes (d) Ours with 200 strokes and $\ell_2$ reward (e) Ours with 200 strokes (Baseline) (f) Ours with 1000 strokes (g) The target images}
\vspace{-0.2cm} 
\label{fig:CelebA}
\end{figure*}

\section{Stroked-based Renderer}
    In this section, we introduce how to build a neural stroke renderer and use it to generate multiple types of strokes.
    
\subsection{Neural Renderer}
    
    Using a neural network to generate strokes has two advantages. First, the neural renderer is flexible to generate any styles of strokes and is more efficient on GPU's than most hand-crafted stroke simulators. Second, the neural renderer is differentiable and enables end-to-end training which boosts the performance of the agent. 
    
    Specifically, the neural renderer has as input a set of stroke parameters $a_t$ and outputs the rendered stroke image $S$. The training samples are generated randomly using Computer Graphics rendering programs. The neural renderer can be quickly trained with supervised learning and runs on the GPU. The model-based transition dynamics $s_{t+1} = \operatorname{trans}(s_t, a_t)$ and the reward function $r(s_t, a_t)$ are differentiable. Some simple geometric trajectories like circles have simple closed-form gradients. However, in general, the discreteness of pixel position and pixel values requires continuous approximation when deriving gradients, e.g.\ for B{\'e}zier Curves. The approximations need to be carefully designed to not break the agent learning.
    
    The neural renderer is a neural network consisting of several fully connected layers and convolution layers. Sub-pixel upsampling~\cite{shi2016real} is used to increase the resolution of strokes in the network, which is a fast running operation and can eliminate the checkerboard effect. We show the network architecture of the neural renderer in Figure~\ref{fig:net_arch} (c).
    
\subsection{Stroke Design}
    Strokes can be designed as a variety of curves or geometries. In general, the parameter of a stroke should include the position, shape, color, and transparency. 
    
    We design a stroke represent of quadratic B{\'e}zier curve (QBC) with thickness to simulate the effects of brushes. The shape of the B{\'e}zier curve is specified by the coordinates of control points. Formally, the stroke is defined as the following tuple:
    \begin{eqnarray}
        a_t = (x_0, y_0, x_1, y_1, x_2, y_2, r_0, t_0, r_1, t_1, R, G, B)_t, 
    \end{eqnarray}
    where $(x_0, y_0, x_1, y_1, x_2, y_2)$ are the coordinates of the three control points of the QBC. $(r0, t0)$, $(r1, t1)$ control the thickness and transparency of the two endpoints of the curve, respectively. $(R, G, B)$ controls the color. The formula of QBC is:
    \begin{eqnarray}
        B(t) = (1-t)^2P_0 + 2(1-t)tP_1 + t^2P_2, 0\leq t \leq 1,
    \end{eqnarray}
    
    
    As changing stroke representation only requires changing the final stroke rendering layer, we can use neural renders with the same network structure to implement the rendering of different stroke designs.
	
\section{Experiments}

    
    Four datasets are used for our experiments, including MNIST~\cite{lecun1998mnist}, SVHN~\cite{netzer2011reading}, CelebA~\cite{liu2015faceattributes} and ImageNet~\cite{ILSVRC15}. We show that the agent has excellent performance in painting various types of real-world images.
    
    \subsection{Datasets}
    MNIST contains 70,000 examples of hand-written digits, where 60,000 are training data, and 10,000 are testing data. Each example is a grayscale image of $28\times28$ pixels.
    
    SVHN is a real-world streetview house number image dataset, including 600,000 digits images. Each sample in the Cropped Digits set is a color image of $32\times32$ pixels. We randomly sample 200,000 images for our experiments.
    
    CelebA contains approximately 200,000 celebrity face images. The officially provided center-cropped images are used in our experiments.
    
    ImageNet (ILSVRC2012) contains 1.2 million natural scene images, which fall into 1000 categories. The extreme diversity of ImageNet poses a grand challenge to the painting agent. We randomly sample 200,000 images that cover 1,000 categories as training data.
    
    In our task, we aim to train an agent that can paint any images rather than only the ones in the training set. Thus, we additionally split out testing set to test the generalization ability of the trained agent. For MNIST, we use the officially defined testing set. For other datasets, we randomly split out 2,000 images as the testing set.
    
    
\begin{figure*}[!t]%
\centering 
\subfigure[DDPG and model-based DDPG]{
\includegraphics[width=0.318\textwidth]{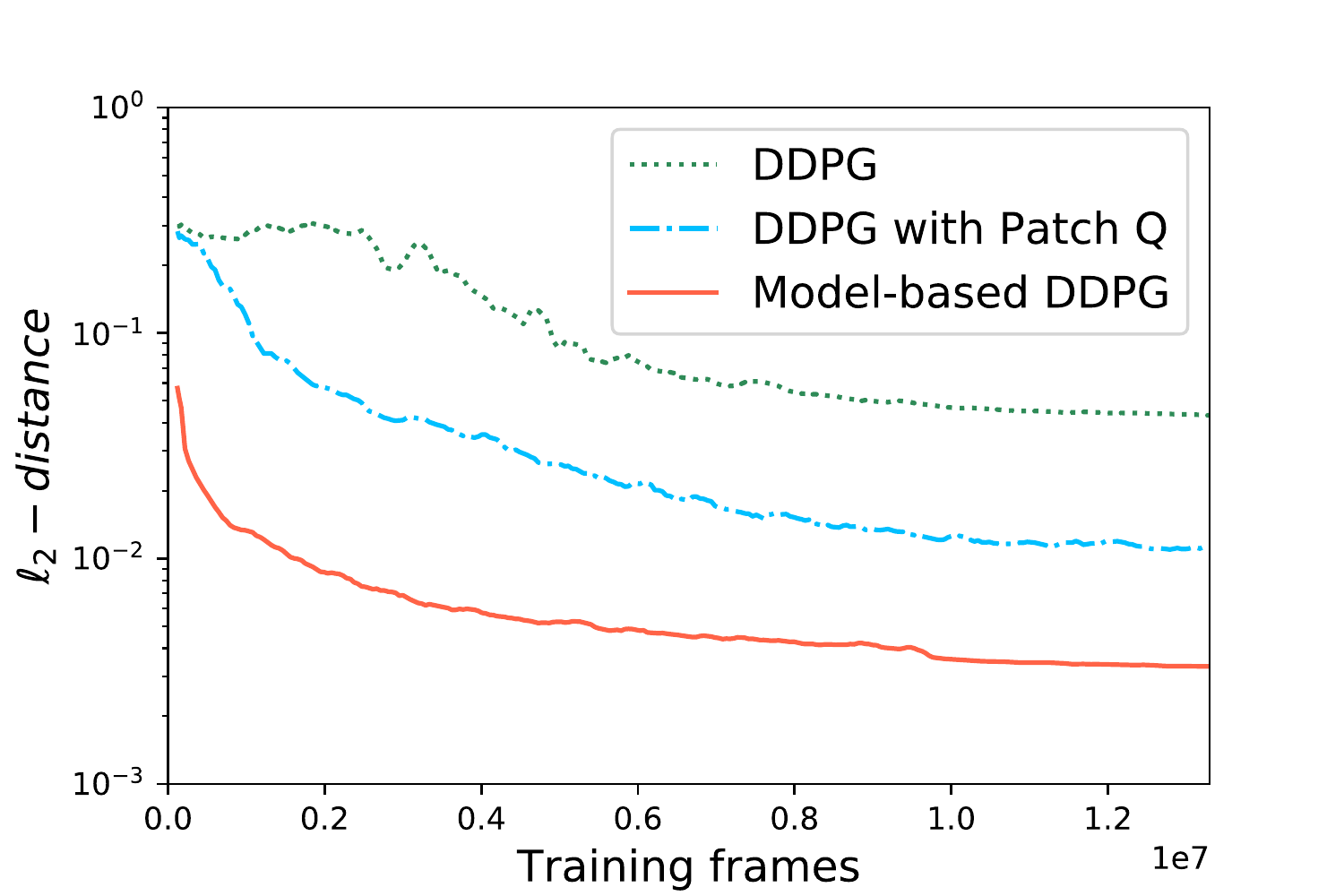}
}
\subfigure[Different settings of Action Bundle]{
\includegraphics[width=0.318\textwidth]{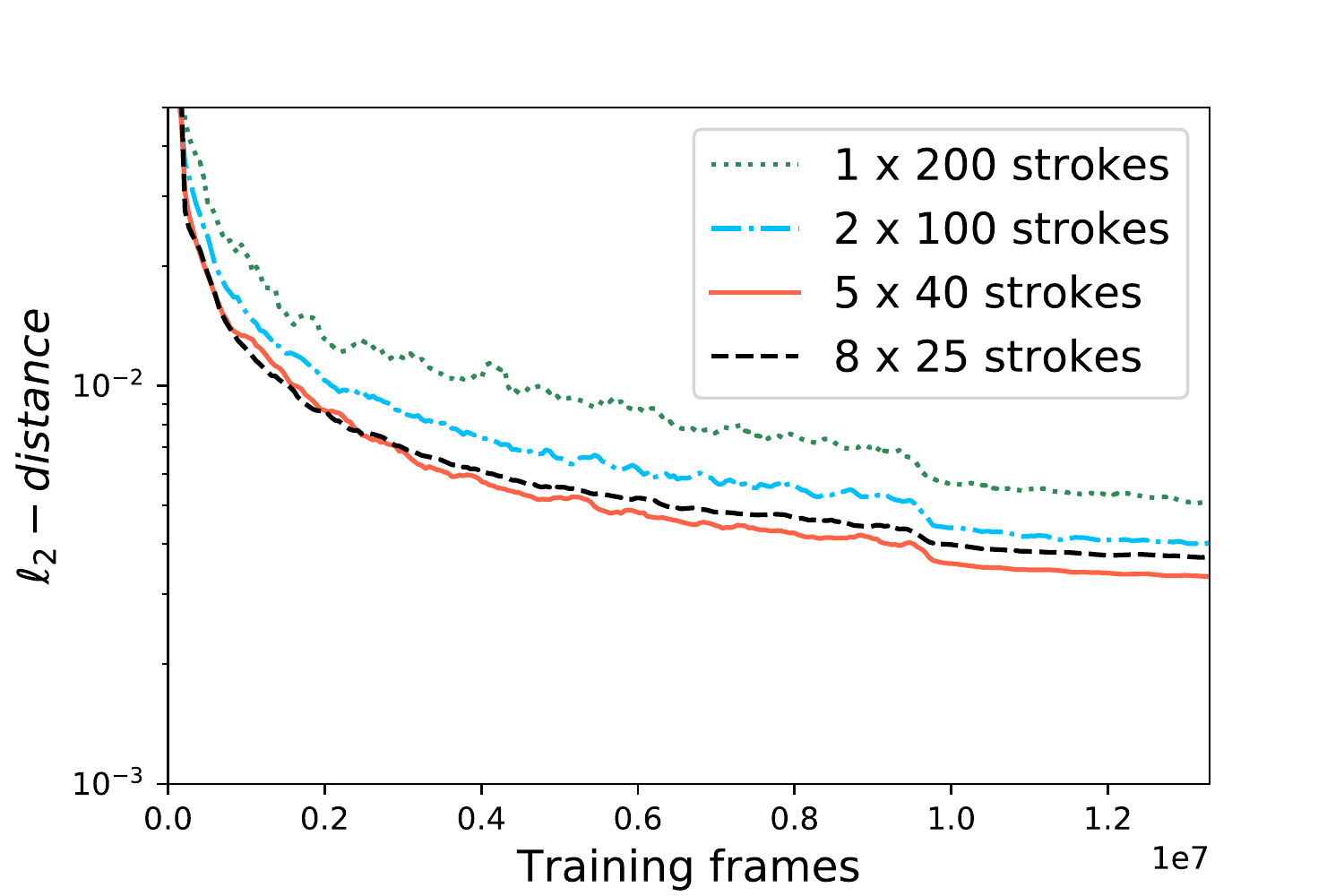}
}
\subfigure[Different number of strokes]{
\includegraphics[width=0.32\textwidth]{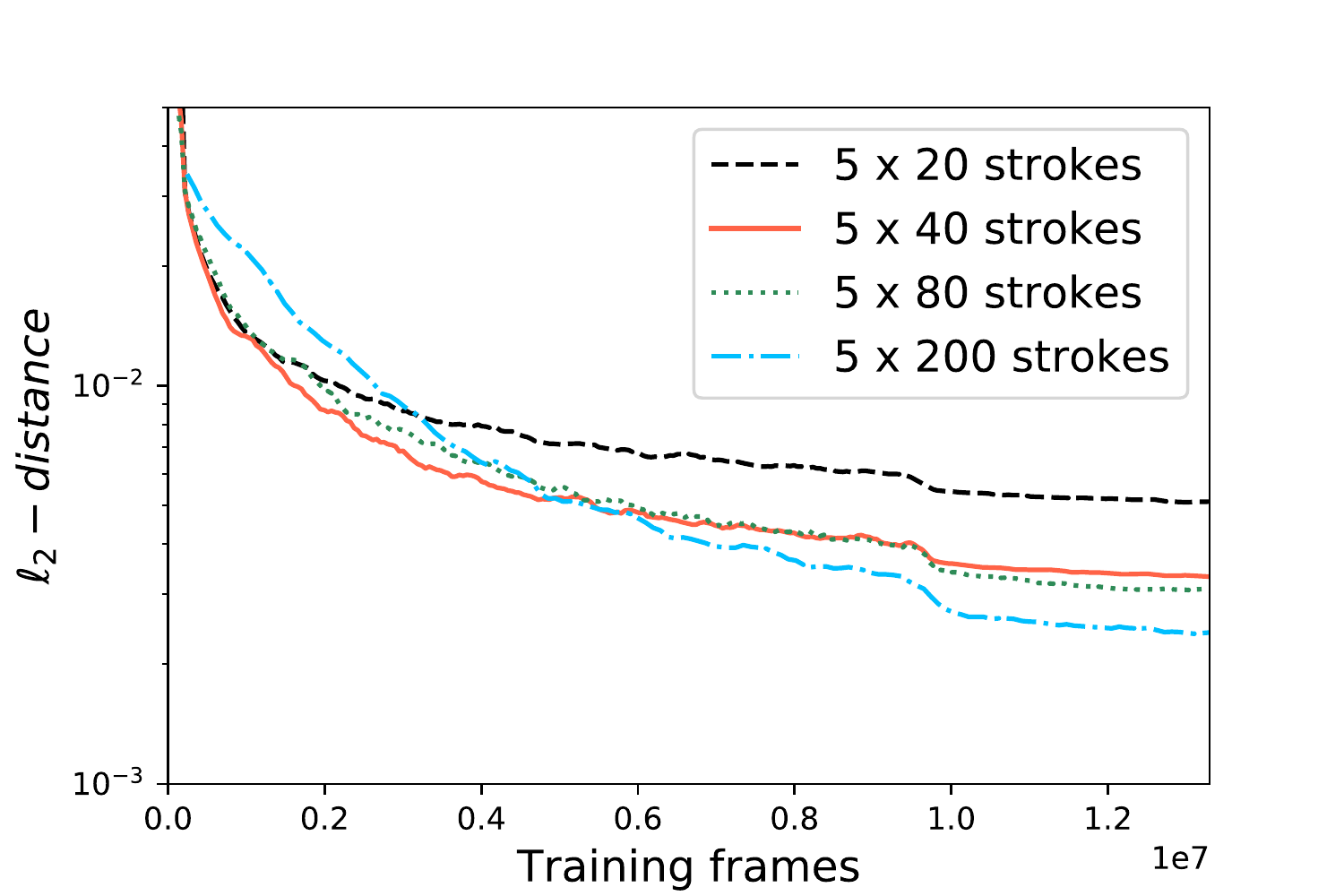}
}
\caption{The testing $\ell_2$-distance between paintings and the target images of CelebA for ablation studies.}
\vspace{-0.2cm}
\label{fig:exp}
\end{figure*}

    \subsection{Training}
    
    We resized all images to the resolution of $128\times128$ pixels before feeding the agent. With an action bundle containing 5 strokes, it takes about $2.1s$ to paint an image using 200 strokes on a 2.2GHz Intel Core i7 CPU. On an NVIDIA 2080Ti GPU, a $9.5\times$ acceleration can be achieved. The computation cost of the actor and renderer are about $554$ MFLOPs and $217$ MFLOPs respectively for painting an action bundle.
    
    We trained the agent with $2\times10^5$ mini-batches for ImageNet and CelebA datasets, $10^5$ mini-batches for SVHN and $2\times10^4$ mini-batches for MNIST. Adam~\cite{kingma2014adam} was used for optimization, and the minibatch size was set as 96. The agent training was performed on a single GPU. It took about 40 hours for training on ImageNet and CelebA, 20 hours for SVHN and two hours for MNIST. It took 5 to 15 hours to train the neural renderer for every stroke design. The same trained renderer can be used for different agents.
    
    At each iteration, we update the critic, actor, and discriminator in turn. All models are trained from scratch. The replay memory buffer was set to store the data of the latest 800 episodes for training the agent. Please refer to the supplemental materials for more training details.
    
    
    
    
    \subsection{Results}
    \begin{figure}[h]%
    \vspace{-0.2cm}
    \centering \includegraphics[width=0.327\textwidth]{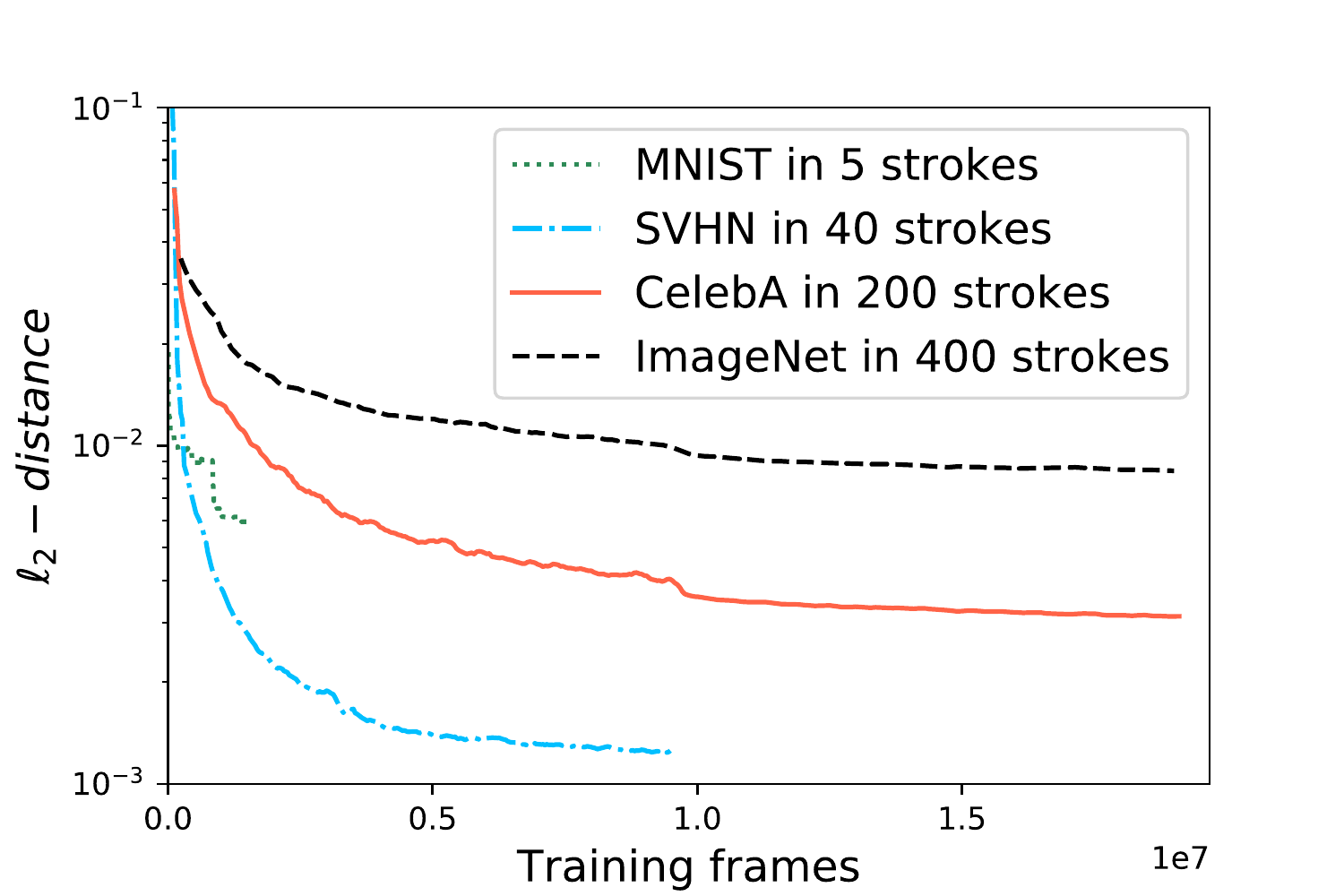}
    \caption{The testing $\ell_2$-distance between the paintings and the target images for different datasets. }
    \label{fig:training_curve}
    \vspace{-0.1cm}
    \end{figure}
    
    The images of MNIST and SVHN show simple image structures and regular contents. We train one agent that paints five strokes for images of MNIST, and another one that paints 40 strokes for images of SVHN. The example paintings are shown in Figure~\ref{fig:4dataset} (a) and (b). The agents can perfectly reproduce the target images.
    
    In contrast, the images of CelebA have more complex structures and diverse contents. We train a 200-strokes agent to deal with the images of CelebA. As shown in Figure~\ref{fig:4dataset} (c), the paintings are quite similar to the target images although losing a certain level of details. 
    
    We train a 400-strokes agent to deal with the images of ImageNet, due to the extremely complex structures and diverse contents. As shown in Figure~\ref{fig:4dataset} (d), paintings are similar to the target images concerning the outline and colors of objects and backgrounds. Despite the loss of some textures, the agent still shows great power in decomposing complicated scenes into strokes and can reasonably repaint them.
    
    We show the test loss curves of agents trained on different datasets in Figure~\ref{fig:training_curve}.
    
    

    
    
    In~\cite{ganin2018synthesizing} SPIRAL shows its performance on CelebA. To make a fair comparison, we also train a 20-strokes agent as SPIRAL and use opaque strokes. The results of the two methods are shown in Figure~\ref{fig:CelebA} (a) and (b) respectively. Our $\ell_2$ distance is 3x smaller than that of SPIRAL. We analyze the main differences between SPIRAL and our methods as follows. First, SPIRAL uses an undifferentiable painting simulator and have to use model-free RL algorithms, which usually perform worse than the model-based ones. Second, SPIRAL predicts an action by recurrently producing each dimension with strong nonlinearities. Our method simplifies this step by predicting multiple action vectors in just one step. Third, we believe having a sufficient number of strokes is critical for vivid results. SPIRAL only gets a reward after finishing a whole episode. This makes the rewards too sparse as the number of steps increases. 
    
    \subsection{Ablation Studies}
    In this section, we study how the components or tricks affect the performance of the agent. The control experiments are performed on CelebA.
    
    \subsubsection{Model-based vs.\ Model-free DDPG}
    We explore how much benefits are brought by model-based DDPG over original DDPG. Original DDPG can only essentially model the environment with observations and rewards from the environment. Besides, the high-dimensional action space also stops model-free methods from successfully dealing with the painting task. To further explore the capability of model-free methods, we improve original DDPG with a method inspired by PatchGAN. We split the images into patches before feeding the critic, then use the patch-level rewards to optimize the critic. We term this method as PatchQ. PatchQ boosts the sample efficiency and improves the performance of the agent by providing much more supervision signals in training.
    
    
    We show the performance of agents trained with different algorithms in Figure~\ref{fig:exp} (a). Model-based DDPG achieves the best performance, with $5\times$ smaller $\ell_2$ distance than DDPG with PatchQ, and $20\times$ smaller $\ell_2$ distance than the original DDPG. Although underperforming the model-based DDPG, DDPG with PatchQ outperforms original DDPG with significant margins.
    
    
    
    \subsubsection{Rewards}
    $\ell_2$ distance is an alternative to the reward for training the actor. We show the painting results of using WGAN rewards (Section ~\ref{sec:wgan_reward}) and $\ell_2$ rewards in Figure~\ref{fig:CelebA} (d) and (e) respectively. The paintings with WGAN rewards show richer textures and look more vivid. Interestingly, we find using WGAN rewards to train the agent can achieve a lower $\ell_2$ loss on the testing data than using $\ell_2$ rewards directly. This shows that WGAN distance is a better metric in measuring the differences between paintings and real-world images than $\ell_2$ distance.
    
    
    \subsubsection{Stroke Number and Action Bundle}
    The stroke number for painting is critical for the final painting quality, especially for texture-rich images. We train agents that can paint 100, 200, 400 and 1000 strokes, and the testing loss curves are shown in Figure~\ref{fig:exp} (c). It's observed that larger stroke numbers contribute to better painting quality. We show the paintings with 200-strokes and 1000-strokes in Figure~\ref{fig:exp} (e) and (f) respectively. To the best of our knowledge, few methods can handle such a large number of strokes. More strokes help reconstruct the details in the paintings.
    
    We show test loss curves of several settings of Action Bundle in Figure~\ref{fig:exp} (b). We find that making the agent predict five strokes in one bundle achieves the best performance. We conjecture that increasing strokes number in one bundle helps the agent to build long-term plans as there will be fewer rounds of decision, even though it will increase the difficulty in a single round of decision. Thus, to achieve a trade-off, a few strokes in one bundle is a good setting for the agent. Experiments determine that setting five actions in an Action Bundle is optimal in our setting for the painting task.
    
    
    \subsubsection{Alternative Stroke Representations}
    
\begin{figure}[htbp]
     \begin{minipage}[t]{0.19\linewidth}
    \vspace{0pt}
    \includegraphics[width=1\linewidth]{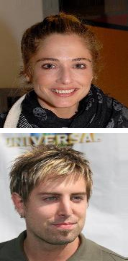}
    \centerline{The target}
    \end{minipage}~
    \begin{minipage}[t]{0.19\linewidth}
    \vspace{0pt}
    \includegraphics[width=1\linewidth]{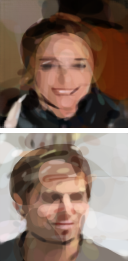}
    \centerline{QBC}
    \end{minipage}
    \begin{minipage}[t]{0.19\linewidth}
    \vspace{0pt}
    \includegraphics[width=1\linewidth]{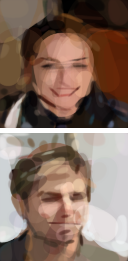}
    \centering
    Straight Stroke
    \end{minipage}
    \begin{minipage}[t]{0.19\linewidth}
    \vspace{0pt}
    \includegraphics[width=1\linewidth]{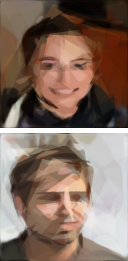}
    \centerline{Triangle}
    \end{minipage}
    \begin{minipage}[t]{0.19\linewidth}
    \vspace{0pt}
    \includegraphics[width=1\linewidth]{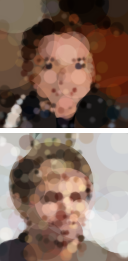}
    \centerline{Circle}
    \end{minipage}
    \caption{CelebA paintings using different stroke designs.}
    \vspace{-0.3cm} 
    \label{fig:diff}
    \end{figure}

    Besides the QBC, we find alternative stroke representations can also be well mastered by the agent, including straight strokes, circles, and triangles. We train one neural renderer for each stroke representation. The paintings with these renderers are shown in Figure~\ref{fig:diff}. The QBC strokes produce excellent visual effects. Meanwhile, other stroke designs create different artistic effects. Although with different styles, the paintings still resemble the target images. This shows that our network architecture generalizes to other choices of stroke designs.
    
    Also, by restricting the transparency of strokes, we can get paintings with different stroke effects, such as ink painting and oil painting as shown in Figure~\ref{fig:CelebA} (c). 
    
	\section{Conclusion}
	In this paper, we train agents that decompose the target image into an ordered sequence of strokes in a fashion mimicking human painting processes on canvases. The training is based on the Deep Reinforcement Learning framework, which encourages the agent to make long-term plans for sequential stroke-based painting. In addition, we build a differentiable neural renderer to render the strokes, which allows using model-based DRL algorithms to further improve the quality of recreated images. The learned agent can predict hundreds or even thousands of strokes to generate a vivid painting. Experimental results show that our model can handle multiple types of target images and achieve good performance in painting real-world images like human portraits and texture-rich natural scenes.
	
	
	

	{\small
		\bibliographystyle{ieee_fullname}    
		\bibliography{egbib}
	}
	\clearpage
	\ificcvfinal\section{Appendix}
\subsection{Architecture}
The network structure diagrams are shown in Figure~\ref{fig:dis},~\ref{fig:renderer},~\ref{fig:actor} and~\ref{fig:critic} , where \texttt{FC} refers to a fully-connected layer, \texttt{Conv} is a convolution layer. The hyperparameters used in training are listed as much as possible in Table~\ref{fig:h1} and~\ref{fig:h2}. All ReLU activations between the layers have been omitted for brevity.

\begin{table}[ht]
	\caption{Hyper-parameters for the \textbf{DDPG}.}
	\begin{tabularx}{\columnwidth}{r|X}
		\toprule
		\# Action per step & $5$ \\
		\# Step per episode & $40$ \\
		Replay buffer size & $800$ episodes\\
		\# Training batches & $2e5$ \\
		Batch size & 96 \\
	    Actor learning rate & \{$3e-4, 1e-4$\} \\
	    Critic learning rate & \{$1e-3, 3e-4$\} \\
	    & * Learning rate decays after $1e5$ training batches \\
	    Reward discount factor & $0.95^5$ \\
		Optimizer & Adam \\
		Actor Normalization & BN \\
		Critic Normalization & WN with TReLU\\
		 \bottomrule
	\end{tabularx}
	\label{fig:h1}
\end{table}
\begin{table}[ht]
	\caption{Hyper-parameters of the \textbf{discriminator} training.}
	\begin{tabularx}{\columnwidth}{r|X}
		\toprule
		Replay buffer size & $800$ episodes\\
		\# Training batches & $2e5$ \\
		Batch size & $96$ \\
		Learning rate & $1e-4$ \\
		Optimizer & Adam ($\beta_1=0.5, \beta_2=0.999$)\\
		Normalization & WN with TReLU \\
		\bottomrule
	\end{tabularx}
	\label{fig:h2}
\end{table}
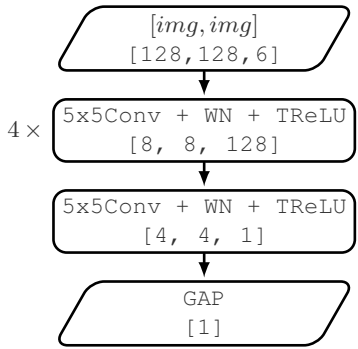
\begin{figure}[ht]
\centering
\begin{tikzpicture}[
  scale=0.3,
  black!100,
  text=black!80,
  node distance=0.35cm and 0.7cm,
  fnode/.style={
    align=center,
    rectangle,minimum height=10pt,minimum width=4cm,rounded corners=5pt,
    inner sep=3pt,
    line width=1pt,
    fill=pnodefill,draw=pnodedraw,
    font=\small\ttfamily},
  nonode/.style={
    align=center,
    font=\small\ttfamily},
  dnode/.style={
    fnode,trapezium,trapezium left angle=70, trapezium right angle=110,trapezium stretches},
  inode/.style={
    dnode,fill=inodefill,draw=inodedraw},
  downnode/.style={
    fnode,trapezium,trapezium left angle=110, trapezium right angle=110,trapezium stretches},
  pnode/.style={
    fnode,fill=pnodefill,draw=pnodedraw},
  pnode2/.style={
    fnode,fill=pnodefill!50!white,draw=pnodedraw!50!white},
  onode/.style={
    dnode,fill=onodefill,draw=onodedraw},
  smnode/.style={
    mnode,fill=mnodefill,draw=mnodedraw},
  flow/.style={
    -latex,shorten >=1pt,line width=1pt,line cap=round,rounded corners=2pt,draw=pnodedraw,draw==\#1},
  flow2/.style={
    flow,draw=pnodedraw!50!white}]
  \newcommand{\mytrap}[3]{%
    \node (#1) [anchor=center,#2] {\phantom{$ State $}\\\phantom{[64,64,1]}}; \node[anchor=center,align=center,font=\small\ttfamily] at (#1.center) {#3};
  }

    \mytrap{input}{dnode}{$ [img, img] $\\{[128,128,6]}};
  \node (conv3) [pnode, below= of input] {5x5Conv + WN + TReLU\\{[8, 8, 128]}};
  \node (manyblocks) [nonode, left=0.0cm of conv3] {$4\,\times$};
  \path (input) edge[flow] (conv3);
  \node (conv4) [pnode, below= of conv3] {5x5Conv + WN + TReLU\\{[4, 4, 1]}};
  \path (conv3) edge[flow] (conv4);
  \mytrap{output}{onode, below= of conv4} {GAP\\{[1]}};
  \path (conv4) edge[flow] (output);
  \end{tikzpicture}
  
\caption{The network architecture of the \textbf{discriminator}. {GAP} refers to a global average pooling layer. Input contains two images.}
\label{fig:dis}
\end{figure}
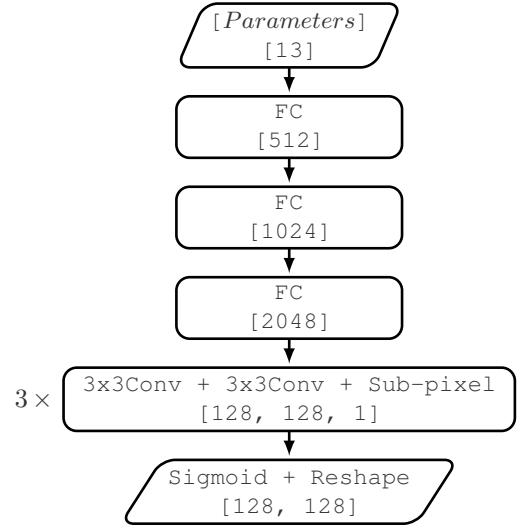
\begin{figure}[ht]
\centering
\begin{tikzpicture}[
  scale=0.3,
  black!100,
  text=black!80,
  node distance=0.35cm and 0.7cm,
  fnode/.style={
    align=center,
    rectangle,minimum height=10pt,minimum width=3cm,rounded corners=5pt,
    inner sep=3pt,
    line width=1pt,
    fill=pnodefill,draw=pnodedraw,
    font=\small\ttfamily},
  nonode/.style={
    align=center,
    font=\small\ttfamily},
  dnode/.style={
    fnode,trapezium,trapezium left angle=70, trapezium right angle=110,trapezium stretches},
  inode/.style={
    dnode,fill=inodefill,draw=inodedraw},
  downnode/.style={
    fnode,trapezium,trapezium left angle=100, trapezium right angle=110,trapezium stretches},
  pnode/.style={
    fnode,fill=pnodefill,draw=pnodedraw},
  pnode2/.style={
    fnode,fill=pnodefill!50!white,draw=pnodedraw!50!white},
  onode/.style={
    dnode,fill=onodefill,draw=onodedraw},
  smnode/.style={
    mnode,fill=mnodefill,draw=mnodedraw},
  flow/.style={
    -latex,shorten >=1pt,line width=1pt,line cap=round,rounded corners=2pt,draw=pnodedraw,draw==\#1},
  flow2/.style={
    flow,draw=pnodedraw!50!white}]
  \newcommand{\mytrap}[3]{%
    \node (#1) [anchor=center,#2] {\phantom{$ State $}\\\phantom{[64,64,1]}}; \node[anchor=center,align=center,font=\small\ttfamily] at (#1.center) {#3};
  }

   \mytrap{input}{dnode}{[$ Parameters $]\\{[13]}};
  \node (fc0) [pnode, below= of input] {FC\\{[512]}};
  \path (input) edge[flow] (fc0);
  \node (fc1) [pnode, below= of fc0] {FC\\{[1024]}};
  \path (fc0) edge[flow] (fc1);
  \node (fc2) [pnode, below= of fc1] {FC\\{[2048]}};
  \path (fc1) edge[flow] (fc2);
  \node (conv) [pnode, below= of fc2, minimum width=6cm] {3x3Conv + 3x3Conv + Sub-pixel \\ {[128, 128, 1]}};
  \path (fc2) edge[flow] (conv);
  \mytrap{output}{onode, below= of conv, minimum width=4.5cm} {Sigmoid + Reshape\\{[128, 128]}};
  \path (conv) edge[flow] (output);
  \node (manyblocks) [left=0.0cm of conv] {$3\,\times$};
  \end{tikzpicture}
  
\caption{The network architecture of the \textbf{neural renderer}.}
\label{fig:renderer}
\end{figure}
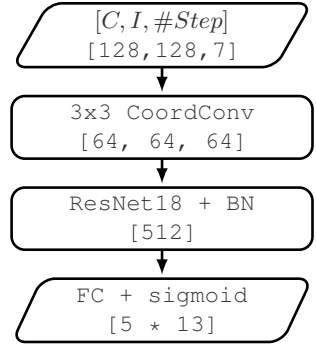
\begin{figure}[ht]
\centering
\begin{tikzpicture}[
  scale=0.3,
  black!100,
  text=black!80,
  node distance=0.35cm and 0.7cm,
  fnode/.style={
    align=center,
    rectangle,minimum height=10pt,minimum width=4cm,rounded corners=5pt,
    inner sep=3pt,
    line width=1pt,
    fill=pnodefill,draw=pnodedraw,
    font=\small\ttfamily},
  nonode/.style={
    align=center,
    font=\small\ttfamily},
  dnode/.style={
    fnode,trapezium,trapezium left angle=70, trapezium right angle=110,trapezium stretches},
  inode/.style={
    dnode,fill=inodefill,draw=inodedraw},
  downnode/.style={
    fnode,trapezium,trapezium left angle=110, trapezium right angle=110,trapezium stretches},
  pnode/.style={
    fnode,fill=pnodefill,draw=pnodedraw},
  pnode2/.style={
    fnode,fill=pnodefill!50!white,draw=pnodedraw!50!white},
  onode/.style={
    dnode,fill=onodefill,draw=onodedraw},
  smnode/.style={
    mnode,fill=mnodefill,draw=mnodedraw},
  flow/.style={
    -latex,shorten >=1pt,line width=1pt,line cap=round,rounded corners=2pt,draw=pnodedraw,draw==\#1},
  flow2/.style={
    flow,draw=pnodedraw!50!white}]
  \newcommand{\mytrap}[3]{%
    \node (#1) [anchor=center,#2] {\phantom{$ State $}\\\phantom{[64,64,1]}}; \node[anchor=center,align=center,font=\small\ttfamily] at (#1.center) {#3};
  }

  \mytrap{input}{dnode}{$ [C, I, \#Step] $\\{[128,128,7]}};
  \node (fc0) [pnode, below= of input] {3x3~CoordConv\\{[64, 64, 64]}};
  \path (input) edge[flow] (fc0);
  \node (fc1) [pnode, below= of fc0] {ResNet18 + BN\\{[512]}};
  \path (fc0) edge[flow] (fc1);
  \node (fc2) [onode, below= of fc1] {FC + sigmoid\\{[5 * 13]}};
  \path (fc1) edge[flow] (fc2);
  \end{tikzpicture}
  
\caption{The network architecture of the \textbf{actor}. Input contains the canvas, target image and step number.}
\label{fig:actor}
\end{figure}
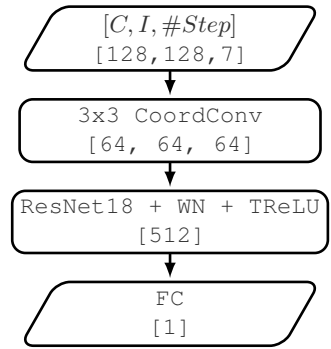
\begin{figure}[ht]
\centering
\begin{tikzpicture}[
  scale=0.3,
  black!100,
  text=black!80,
  node distance=0.35cm and 0.7cm,
  fnode/.style={
    align=center,
    rectangle,minimum height=10pt,minimum width=4cm,rounded corners=5pt,
    inner sep=3pt,
    line width=1pt,
    fill=pnodefill,draw=pnodedraw,
    font=\small\ttfamily},
  nonode/.style={
    align=center,
    font=\small\ttfamily},
  dnode/.style={
    fnode,trapezium,trapezium left angle=70, trapezium right angle=110,trapezium stretches},
  inode/.style={
    dnode,fill=inodefill,draw=inodedraw},
  downnode/.style={
    fnode,trapezium,trapezium left angle=110, trapezium right angle=110,trapezium stretches},
  pnode/.style={
    fnode,fill=pnodefill,draw=pnodedraw},
  pnode2/.style={
    fnode,fill=pnodefill!50!white,draw=pnodedraw!50!white},
  onode/.style={
    dnode,fill=onodefill,draw=onodedraw},
  smnode/.style={
    mnode,fill=mnodefill,draw=mnodedraw},
  flow/.style={
    -latex,shorten >=1pt,line width=1pt,line cap=round,rounded corners=2pt,draw=pnodedraw,draw==\#1},
  flow2/.style={
    flow,draw=pnodedraw!50!white}]
  \newcommand{\mytrap}[3]{%
    \node (#1) [anchor=center,#2] {\phantom{$ State $}\\\phantom{[64,64,1]}}; \node[anchor=center,align=center,font=\small\ttfamily] at (#1.center) {#3};
  }

  \mytrap{input}{dnode}{$ [C, I, \#Step] $\\{[128,128,7]}};
  \node (fc0) [pnode, below= of input] {3x3~CoordConv\\{[64, 64, 64]}};
  \path (input) edge[flow] (fc0);
  \node (fc1) [pnode, below= of fc0] {ResNet18 + WN + TReLU\\{[512]}};
  \path (fc0) edge[flow] (fc1);
  \mytrap{output}{onode, below= of fc1} {FC\\{[1]}};
  \path (fc1) edge[flow] (output);
  \end{tikzpicture}
  
\caption{The network architecture of the \textbf{critic}. Input contains the canvas, target image and step number.}
\label{fig:critic}
\end{figure} \fi
\end{document}